\documentclass{article}

 \usepackage[preprint]{neurips_2026}

\usepackage[utf8]{inputenc} % allow utf-8 input
\usepackage[T1]{fontenc}    % use 8-bit T1 fonts
\usepackage{hyperref}       % hyperlinks
\usepackage{url}            % simple URL typesetting
\usepackage{booktabs}       % professional-quality tables
\usepackage{amsfonts}       % blackboard math symbols
\usepackage{nicefrac}       % compact symbols for 1/2, etc.
\usepackage{microtype}      % microtypography
\usepackage[table]{xcolor}  % colors and table row shading

\usepackage{graphicx}
\usepackage{subcaption}
\usepackage{amsmath}
\usepackage{bm}
\usepackage{algorithm}
\usepackage{algorithmic}
\usepackage{multirow}

\title{Federated Nested Learning: Collaborative Training of Self-Referential Memories for Test-Time Adaptation}

\author{%
  Hong Chen\thanks{Co-first authors.} \\
  HKUST (GZ) \\
  \texttt{hchen763@connect.hkust-gz.edu.cn} \\
  \AND
  Pengcheng Wu\footnotemark[1] \\
  Nanyang Technological University \\
  \texttt{pengchengwu@ntu.edu.sg} \\
  \And
  Yuanguo Lin\thanks{Corresponding author.} \\
  Jimei University \\
  \texttt{xdlyg@jmu.edu.cn} \\
  \AND
  Peilin Zhao \\
  Shanghai Jiao Tong University \\
  \texttt{peilinzhao@sjtu.edu.cn} \\
  \AND
  Xiuze Zhou \\
  HKUST (GZ) \\
  \texttt{xzhou154@connect.hkust-gz.edu.cn} \\
  \And
  Fan Lin \\
  Xiamen University \\
  \texttt{iamafan@xmu.edu.cn} \\
  \AND
  Han Yu \\
  Nanyang Technological University \\
  \texttt{han.yu@ntu.edu.sg} \\
}

\begin{document}

\maketitle

\begin{abstract}
    We rethink Federated Learning (FL) from a nested learning perspective, framing the core challenge as how to collaboratively learn optimization rules, not just static models, to tackle Non-IID client data. To address this, we propose Federated Nested Learning (FedNL), a novel framework that reformulates FL as a three-level nested optimization system. FedNL embeds Titans-based linear attention into FL, enabling clients to perform lightweight, zero-shot test-time adaptation by treating a delta rule as an online gradient step. Experiments on Non-IID MMLU and long-context benchmarks show that FedNL achieves competitive performance in short-context reasoning, enhances the performance of long-context retrieval and streaming Cross-Entropy, and maintains constant inference memory.
\end{abstract}

\section{Introduction}
Federated Learning (FL) has emerged as a privacy-preserving paradigm for collaboratively training large language models (LLMs) across distributed edge devices~\citep{kuang2024federatedscope, ye2024openfedllm}. By keeping raw data local and aggregating model updates, FL promises to harness the collective intelligence of massive, decentralized datasets. However, the real-world deployment of Federated LLMs faces two persistent and intertwined challenges: {data heterogeneity (Non-IID)} and {long-tail distributions}. In realistic scenarios, client data distributions are highly skewed (e.g., a medical client vs. a coding assistant), and critical knowledge often resides in the long tail of these distributions, which is easily overshadowed by head classes during global aggregation~\citep{shuai2022balancefl}.

To address the above challenges, existing approaches primarily focus on \textit{regularizing} or \textit{personalizing} the static model weights. For instance, FedProx~\citep{li2020federated} introduces proximal terms to restrict local deviation, while recent state-of-the-art methods like FedSSI~\citep{li2025fedssi} employ synaptic intelligence to selectively preserve important parameters, effectively mitigating catastrophic forgetting. Despite their success, these methods share a fundamental limitation: they treat the global model as a container of {static knowledge}. When such a static model is deployed to a client with unseen, highly heterogeneous data, it lacks \textit{test-time plasticity} --- the ability to adapt to the current context without gradient updates. Consequently, static weights often fail to capture the nuances of long-tail distributions that are context-dependent, leading to suboptimal performance on domain-specific tasks~\citep{wang2023dafkd}.

In this paper, we argue that solving the Non-IID and long-tail dilemma requires a paradigm shift from aggregating static knowledge to aggregating {learning capabilities}. Drawing inspiration from the emerging theory of {Nested Learning (NL)}~\citep{behrouz2025nested}, we posit that learning should not be dichotomized into ``training'' and ``inference'', but rather viewed as a hierarchy of nested optimization processes operating at different frequencies. From this perspective, the ``inference'' phase of a sequence model can be reframed as a high-frequency ``inner-loop training'' process, where the model actively compresses the current context into a transient memory state. If a model possesses a powerful mechanism to construct this memory at test time, it can dynamically adapt to heterogeneous local distributions without altering its global weights.

Building on this insight, we propose {Federated Nested Learning (FedNL)}, a novel framework that reformulates FL as a three-level nested optimization system. Instead of aggregating the memory content itself (which is private and heterogeneous), FedNL aggregates the {meta-rules} governing how memory is constructed and updated. Specifically, we leverage the {Titans} architecture~\citep{behrouz2024titans}, which utilizes a linearized attention mechanism equipped with a \textit{Delta Rule}. In our framework, the server aggregates the projection matrices and gating coefficients (Level 0), while clients utilize these global rules to instantiate private, context-aware memory states $S_t$ during local inference (Level 2). This {self-referential} mechanism allows the model to ``learn to memorize'' the specific patterns of local long-tail data on-the-fly, effectively bypassing the limitations of static weight aggregation.

Our approach offers a parameter-efficient way to address aspects of traditional FL heterogeneity. By decoupling general linguistic capabilities (frozen backbone) from memory construction rules (trainable adapters), FedNL achieves superior adaptability with minimal communication overhead. We implement FedNL using the computationally efficient LiZAttention module~\citep{furfaro2025tptt} and validate it on diverse benchmarks.
Our main contributions are summarized as follows.
\begin{itemize}
    \item \textbf{Theoretical Reframing:} We introduce the Nested Learning perspective to FL, formalizing the problem as a collaborative training of optimization rules rather than static representations. This provides a theoretical basis for addressing Non-IID issues via test-time adaptation.
    \item \textbf{The FedNL Framework:} We propose a practical algorithm that integrates Titans-based linear attention into the FL pipeline. By treating the Delta Rule as an online gradient descent step, we enable clients to perform \textit{Zero-Shot Test-Time Adaptation} on unseen domains without computational heavy lifting.
    \item \textbf{Empirical Superiority:} Experiments on Non-IID MMLU and long-context benchmarks show that FedNL is competitive with strong federated baselines on short-context reasoning and obtains larger gains on long-context retrieval and streaming Cross-Entropy (CE) diagnostics. Notably, a 16K Needle In A Haystack (NIAH) streaming CE probe shows that FedNL continues to reduce normalized loss as context unfolds while FedAvg accumulates uncertainty, and does so while maintaining constant inference memory complexity.
\end{itemize}

%%%%%%%%%%%%%%%%%%%%%%%%%%%%%%%%%%%%%%%%%%%%%%%%%%%%%%%%%%%%%%%%%%%%%%%%%%%%%%%

%%%%%%%%%%%%%%%%%%%%%%%%%%%%%%%%%%%%%%%%%%%%%%%%%%%%%%%%%%%%%%%%%%%%%%%%%%%%%%%
\section{Methodology}
\label{sec:methodology}

\subsection{Preliminaries}
\label{sec:preliminaries}

In this section, we formalize the problem of FL with Test-Time Adaptation constraints. We then review the formulation of Linear Attention mechanisms (specifically Titans) as associative memory optimization. Finally, drawing on NL theory, we formally define our proposed {federated nested optimization} framework.

\textbf{Federated Learning with Test-Time Adaptation.}

Consider a federated learning system with $K$ clients, where each client $k$ holds a private dataset $\mathcal{D}_k = \{(x^{(i)}, y^{(i)})\}_{i=1}^{N_k}$ drawn from a local distribution $\mathcal{P}_k$. The standard goal of FL is to find a global parameter vector $\theta^*$ that minimizes the weighted empirical risk over all clients:
\begin{equation}
    \theta^* = \arg\min_{\theta} \sum_{k=1}^K \frac{N_k}{N} \mathcal{L}_k(\theta; \mathcal{D}_k),
\end{equation}
where $\mathcal{L}_k$ is the local loss function (e.g., Cross-Entropy) and $N = \sum N_k$.

\textbf{The Challenge of Static Weights.} In traditional settings, once $\theta^*$ is deployed to a client $k$ for inference (test-time), the parameters remain fixed. Let $x_{1:T} = (x_1, \dots, x_T)$ be a test sequence on client $k$. A static model computes predictions $p(x_{t+1}|x_{1:t}; \theta^*)$. If the test distribution $\mathcal{P}_{test}$ significantly shifts from the training distributions (i.e., extreme Non-IID or Long-Tail scenarios), the static $\theta^*$ struggles to adapt.

\textbf{Test-Time Adaptation (TTA).} To address this, we consider a setting where the model maintains a dynamic state $S_t$ during inference. The prediction becomes $p(x_{t+1}|x_{1:t}; S_t, \theta)$, where $S_t$ is updated online based on the context $x_{1:t}$. Our goal in {FedNL} is to learn the optimal \textit{update rules} (encoded in $\theta$) such that $S_t$ rapidly converges to a representation that minimizes local prediction error at test time, without requiring gradient updates to $\theta$ itself.

\textbf{Neural Memory as Online Optimization.}

We leverage the {Titans} architecture~\citep{behrouz2024titans}, which treats the attention mechanism as a Neural Memory module. Unlike standard Softmax attention which requires storing the full history buffer, Titans compresses history into a fixed-size memory state $\mathbf{S} \in \mathbb{R}^{d \times d}$.

From the NL perspective~\citep{behrouz2025nested}, the update of this memory state is not merely a heuristic recurrence, but an online optimization step. Specifically, let $\mathbf{k}_t, \mathbf{v}_t \in \mathbb{R}^d$ be the key and value vectors projected from input $x_t$ using parameters $\theta$. The memory state $\mathbf{S}_t$ is updated to map keys to values by minimizing a momentary associative memory objective:
\begin{equation}
    \mathbf{S}_t = \arg\min_{\mathbf{S}} \left( \frac{1}{2} \| \mathbf{S}\mathbf{k}_t - \mathbf{v}_t \|^2 + \frac{1}{2\eta} \| \mathbf{S} - \mathbf{S}_{t-1} \|^2 \right).
    \label{eq:inner_objective}
\end{equation}
Solving Eq. (\ref{eq:inner_objective}) via one step of Gradient Descent yields the \textbf{Delta Rule} update:
\begin{equation}
    \mathbf{S}_t = \mathbf{S}_{t-1} - \eta \nabla_{\mathbf{S}} \mathcal{L}_{mem} = \mathbf{S}_{t-1} + \eta (\mathbf{v}_t - \mathbf{S}_{t-1}\mathbf{k}_t)\mathbf{k}_t^\top,
    \label{eq:delta_rule}
\end{equation}
where $\eta$ is a learnable step size (or gating factor) derived from $\theta$. This formulation reveals that \textit{inference is effectively a high-frequency training process}, where the model ``learns'' the current context by optimizing $\mathbf{S}_t$.

\textbf{The Federated Nested Optimization Framework.}

Building on the concepts above, we formalize FedNL as a three-level nested optimization problem. This framework decouples the global learning of rules from the local construction of memory. We define the system tuple $\mathcal{N} = \{ (L_0, \mathcal{T}_0), (L_1, \mathcal{T}_1), (L_2, \mathcal{T}_2) \}$, representing the three levels of optimization loops:

\textbf{(1) The Inner Loop: Test-Time Adaptation (Client-Side).}
This loop runs during inference on the client. It is ``unsupervised'' in the sense that it does not require ground-truth labels $y$, but self-supervised by the memory objective.
For a given client $k$ and input stream $x_{1:T}$, the memory state trajectory $\mathcal{S}_k = (\mathbf{S}_0, \dots, \mathbf{S}_T)$ is generated by recursively solving the inner objective defined in Eq. (\ref{eq:inner_objective}):
\begin{equation}
    \mathbf{S}_t(\theta) = \text{OnlineOptimizer}(\mathbf{S}_{t-1}; \theta, x_t),
    \label{eq:inner_loop}
\end{equation}
where $\text{OnlineOptimizer}$ corresponds to the Delta Rule update. Note that $\mathbf{S}_t$ is strictly a function of the local context and the parameters $\theta$. This state is transient and private, never leaving the device.

\textbf{(2) The Intermediate Loop: Rule Learning (Client-Side).}
This loop runs during the local training phase. The client optimizes the parameters $\theta$ (e.g., LoRA weights, gating mechanisms) to ensure that the Inner Loop produces a memory state $\mathbf{S}_t$ that is useful for the downstream task (e.g., next-token prediction).
The local objective for client $k$ is:
\begin{equation}
    \min_{\theta} \mathcal{J}_k(\theta) = \mathbb{E}_{(x, y) \sim \mathcal{D}_k} \left[ \sum_{t} \mathcal{L}_{\text{task}}(f(x_t, \mathbf{S}_{t-1}(\theta)), y_t) \right],
    \label{eq:intermediate_loop}
\end{equation}
where $f$ is the prediction head. Crucially, calculating the gradient $\nabla_\theta \mathcal{J}_k$ requires differentiating \textit{through} the Inner Loop process (Eq. \ref{eq:inner_loop}), a technique known as Backpropagation Through Time (BPTT) in RNNs or Meta-Gradients in meta-learning. This ensures $\theta$ learns \textit{how to construct memory} for the specific data distribution of client $k$.

\textbf{(3) The Outer Loop: Collaborative Generalization (Server-Side).}
This loop runs on the server to aggregate the locally learned rules. Since $\theta$ represents the ``physics'' of memory construction rather than the memory itself, aggregating $\theta$ allows diverse clients to share learning capabilities.
The global objective is:
\begin{equation}
    \min_{\theta} \mathcal{G}(\theta) = \sum_{k=1}^K \frac{N_k}{N} \mathcal{J}_k(\theta).
\end{equation}
The server performs the update $\theta^{r+1} \leftarrow \text{Aggregate}(\{\theta_k^{r+1}\}_{k})$, typically via weighted averaging (FedAvg).

\textbf{Unified View.} By nesting these loops, FedNL effectively trains a distributed \textit{optimizer}. The global model $\theta$ is not a static knowledge base, but a \textit{meta-learner}. When deployed to a new client with a Non-IID distribution (e.g., medical records), the meta-learner $\theta$ executes the Inner Loop to rapidly build a medical-specific memory $\mathbf{S}_t$ from the context, achieving zero-shot adaptation without explicit gradient updates.

\textit{Detailed derivations of the gradient flow through the memory states and the implementation of efficient chunk-wise parallelization are provided in Appendix A.}

%%%%%%%%%%%%%%%%%%%%%%%%%%%%%%%%%%%%%%%%%%%%%%%%%%%%%%%%%%%%%%%%%%%%%%%%%%%%%%%
% \begin{figure}[htbp]
%     \centering
%     \includegraphics[width=0.9\linewidth]{Framework.pdf}
%     \caption{Caption}
%     \label{fig:placeholder}
% \end{figure}

\begin{figure}[t]
    \centering
    \begin{minipage}[t]{0.48\textwidth}
        \centering
        \includegraphics[width=\linewidth]{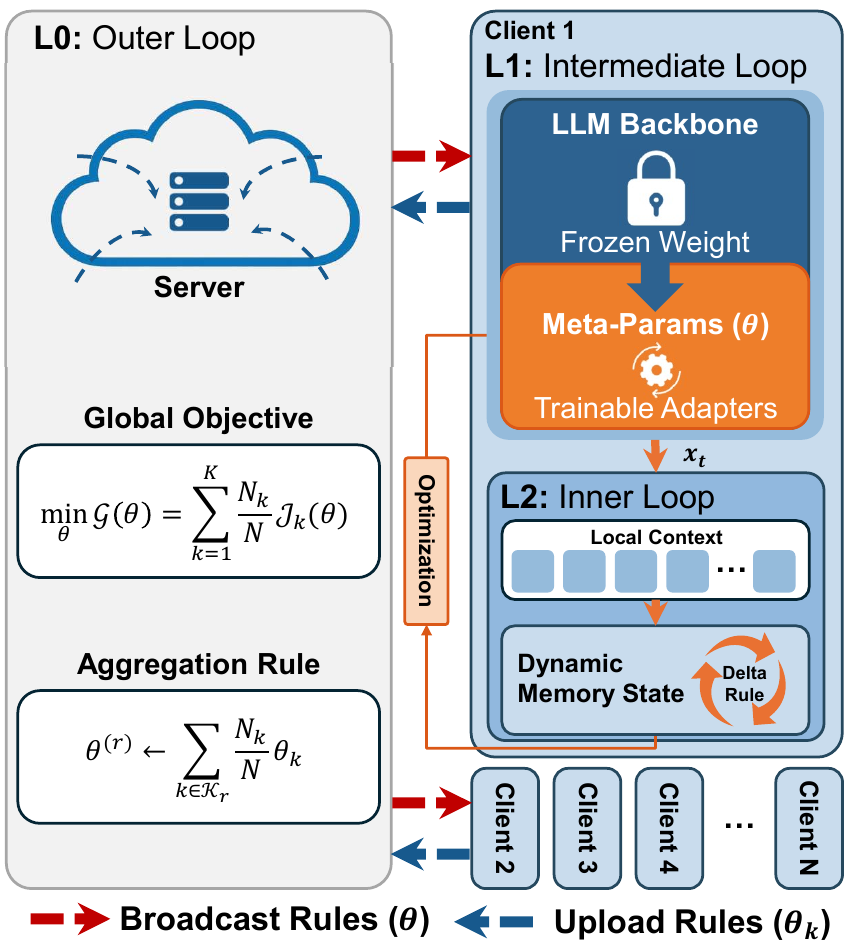}
        \caption{The three-level nested optimization framework of FedNL. L2: Memory state $S_t$ updated via the Delta Rule for test-time adaptation. L1: Meta-parameters $\theta$ (LoRA adapters) trained with frozen backbone. L0: Server aggregates rules $\theta$, not private memory. Red: parameter flow; Blue: meta-gradient flow.}
        \label{fig:framework}
    \end{minipage}
    \hfill
    \begin{minipage}[t]{0.48\textwidth}
        \centering
        \includegraphics[width=\linewidth]{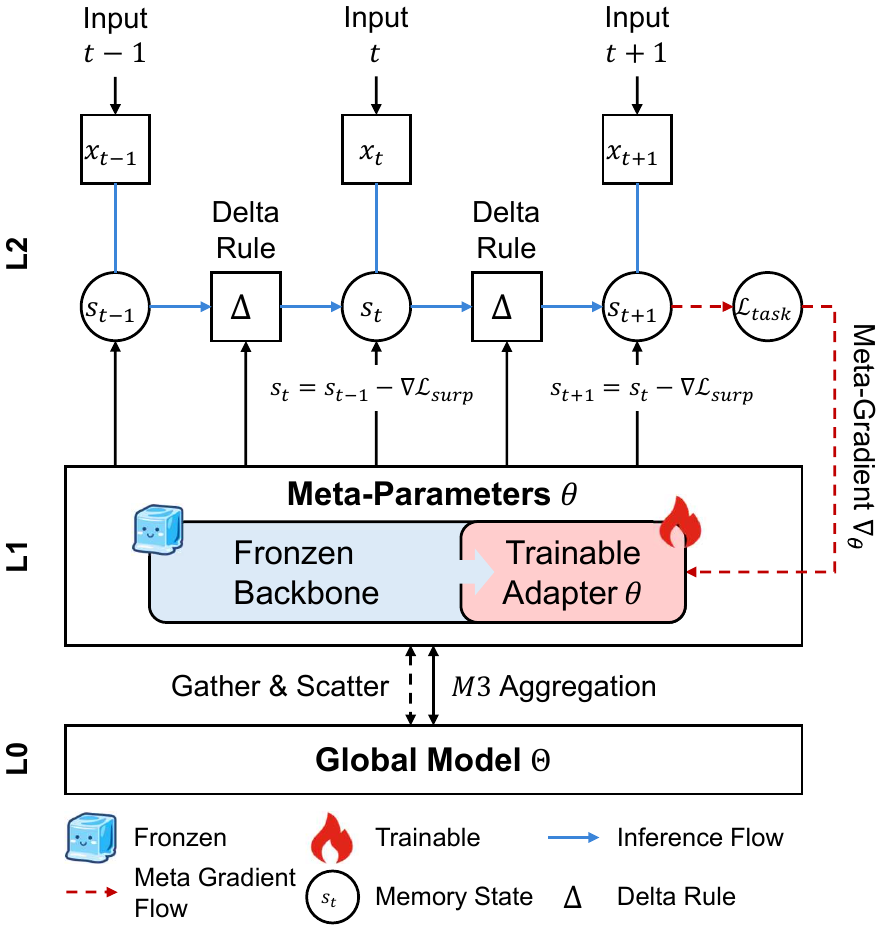}
        \caption{Unrolled computation graph of FedNL. L2: Token-level memory updates $s_t = s_{t-1} - \nabla\mathcal{L}_{\text{surp}}$ via Delta Rule. L1: Meta-gradients $\nabla_\theta$ backpropagated through memory trajectory. L0: M3 aggregation of global meta-rules $\theta$. Memory states remain strictly local.}
        \label{fig:algorithm}
    \end{minipage}
\end{figure}

Based on the theoretical framework established in Section \ref{sec:preliminaries}, we present the implementation of \textbf{FedNL}. We first detail the model architecture that decouples static linguistic capabilities from dynamic memory rules. We then describe the training algorithm that coordinates the three-level nested optimization. Finally, we provide an analytical understanding of why this self-referential mechanism is inherently robust to data heterogeneity.

\subsection{Architecture: Decoupling Knowledge and Rules}
To implement FedNL efficiently under resource-constrained settings, we build upon the {LiZAttention} mechanism~\citep{furfaro2025tptt}, which integrates linear attention into pretrained Transformers. We define the global model $\mathcal{M}$ as a composition of three distinct components:

\textbf{1. The Frozen Backbone ($\Theta_{\text{fixed}}$):} We utilize a pretrained Large Language Model (e.g., Llama-3.2-1B) as the backbone. All original weights (Self-Attention, FFN, Norms) remain frozen throughout the federated lifecycle. This component provides general linguistic knowledge and feature extraction capabilities, acting as a shared basis across all clients.

\textbf{2. The Dynamic Memory Module ($\mathbf{S}_t$):} We replace standard Softmax Attention with a dual-path mechanism. The \textit{Linear Path} maintains the transient memory state $\mathbf{S}_t$ updated via the Delta Rule (Eq. \ref{eq:delta_rule}). This state acts as a private, context-specific container that captures local distribution patterns during inference.

\textbf{3. The Trainable Meta-Parameters ($\theta$):} These are the only parameters communicated and updated in FedNL. They consist of:
\begin{itemize}
    \item \textbf{Low-Rank Projections (LoRA):} We inject low-rank matrices $A, B$ into the query, key, and value projections: $W' = W_{\text{fixed}} + BA$. These learnable adapters determine \textit{what} information should be written into the memory $\mathbf{S}_t$ and \textit{how} it should be retrieved.
    \item \textbf{Memory Gating ($\alpha$):} A learnable scalar or vector that controls the mixing weight between the static Softmax attention (general knowledge) and dynamic Linear attention (local context memory).
\end{itemize}

By restricting the learnable parameters $\theta$ to the adapters, FedNL reduces communication overhead by orders of magnitude compared to full-model aggregation, while the dynamic $\mathbf{S}_t$ provides infinite capacity for test-time context compression.

\subsection{The FedNL Algorithm}

% \begin{figure}[t]
%     \centering
%     \includegraphics[width=0.6\linewidth]{Algorithm.pdf}
%     \caption{Caption}
%     \label{fig:placeholder}
% \end{figure}

The training procedure of FedNL simulates the nested learning process. The core innovation lies in the client's local update step, where the gradient calculation must account for the trajectory of the dynamic memory $\mathbf{S}_t$.

\textbf{Forward Pass (Inner Loop Execution):}
During local training on a sequence $x$, the client executes the model forward pass. Crucially, this is not just a function evaluation but an optimization process. For each token step $t$, the Delta Rule updates $\mathbf{S}_{t-1} \to \mathbf{S}_t$ using the current rules $\theta$. The prediction $\hat{y}_{t+1}$ depends on $\mathbf{S}_t$, which in turn depends on the history $x_{1:t}$ and $\theta$.

\textbf{Backward Pass (Rule Optimization):}
To optimize $\theta$, we compute the gradient of the task loss $\mathcal{L}_{\text{task}}$. Since $\mathbf{S}_t$ is a function of $\theta$ (recursively), the gradient flows through time:
\begin{equation}
    \frac{\partial \mathcal{L}_{\text{task}}}{\partial \theta} = \sum_{t} \frac{\partial \mathcal{L}_t}{\partial \hat{y}_t} \left( \frac{\partial \hat{y}_t}{\partial \theta} + \frac{\partial \hat{y}_t}{\partial \mathbf{S}_{t-1}} \underbrace{\frac{\partial \mathbf{S}_{t-1}}{\partial \theta}}_{\text{Recursive Term}} \right).
    \label{eq:gradient_flow}
\end{equation}
Modern automatic differentiation frameworks handle this BPTT (Backpropagation Through Time) naturally. By minimizing this loss, $\theta$ learns to generate update rules that maximize the predictive power of the memory $\mathbf{S}_t$. The full procedure is detailed in Algorithm \ref{alg:fednl}.

\begin{algorithm}[t]
\caption{Federated Nested Learning (FedNL)}
\label{alg:fednl}
\begin{algorithmic}[1]
\STATE \textbf{Input:} Pretrained Backbone $\Theta_{\text{fixed}}$, Clients $K$, Rounds $R$, Local Epochs $E$.
\STATE \textbf{Server Initialize:} Meta-parameters $\theta^{(0)}$ (LoRA + Gating).
\FOR{round $r = 1$ to $R$}
    \STATE Server selects subset of clients $\mathcal{K}_r$.
    \STATE Broadcast $\theta^{(r-1)}$ to clients in $\mathcal{K}_r$.
    \FOR{client $k \in \mathcal{K}_r$ \textbf{in parallel}}
        \STATE $\theta_k \leftarrow \theta^{(r-1)}$
        \FOR{epoch $e = 1$ to $E$}
            \FOR{batch $B = (x, y)$ in $\mathcal{D}_k$}
                \STATE Initialize memory state $\mathbf{S}_0 = \mathbf{0}$.
                \STATE \textbf{// Level 2 Loop (Implicit)}
                \FOR{token $t$ in sequence}
                    \STATE Generate $k_t, v_t, q_t$ using $\Theta_{\text{fixed}} + \theta_k$.
                    \STATE Update $\mathbf{S}_t \leftarrow \mathbf{S}_{t-1} + \text{DeltaRule}(k_t, v_t)$.
                    \STATE Compute output using $\mathbf{S}_t$.
                \ENDFOR
                \STATE Compute Loss $\mathcal{L} = \text{CrossEntropy}(\text{output}, y)$.
                \STATE \textbf{// Level 1 Loop}
                \STATE Update $\theta_k \leftarrow \theta_k - \eta \nabla_{\theta_k} \mathcal{L}$.
            \ENDFOR
        \ENDFOR
        \STATE Return $\theta_k$ to Server.
    \ENDFOR
    \STATE \textbf{// Level 0 Loop}
    \STATE $\theta^{(r)} \leftarrow \sum_{k \in \mathcal{K}_r} \frac{N_k}{N} \theta_k$.
\ENDFOR
\end{algorithmic}
\end{algorithm}

\subsection{Theoretical Analysis}
\label{subsec:analysis}

Standard FL fails in Non-IID settings because it tries to find a single static parameter set $\theta^*$ that satisfies conflicting local distributions. Specifically, let $\mathcal{P}_1$ and $\mathcal{P}_2$ be two disparate distributions (e.g., Code vs. Medical). A static model attempts to find $\theta^* \in \arg\min (\mathcal{L}_{\mathcal{P}_1}(\theta) + \mathcal{L}_{\mathcal{P}_2}(\theta))$, often resulting in a solution that is suboptimal for both (the ``average'' model).

In FedNL, the prediction for a sample $x$ is not determined by $\theta$ alone, but by the tuple $(\theta, \mathbf{S}_x)$, where $\mathbf{S}_x$ is the memory state dynamically constructed from the context of $x$ itself.

\textbf{Proposition 1 (Instance-Specific Approximation).}
Let $\theta^*$ be the aggregated meta-parameters in FedNL. For any client $k$ with distribution $\mathcal{P}_k$, and for any instance $x \sim \mathcal{P}_k$, the effective model used for prediction is $\mathcal{M}(x; \theta^*) \approx \mathcal{M}_{static}(\theta^* + \Delta \theta_x)$, where $\Delta \theta_x$ represents an implicit gradient step taken on the memory state $\mathbf{S}$ during inference.

\textit{Proof Sketch.} The Delta Rule update $\mathbf{S}_t = \mathbf{S}_{t-1} - \eta \nabla \mathcal{L}_{mem}$ in Level 2 is mathematically equivalent to a gradient descent step in the function space of linear layers~\citep{von2023transformers, behrouz2025nested}. Therefore, when FedNL processes a medical text, the memory $\mathbf{S}$ moves in the direction that minimizes the reconstruction error of medical tokens. This is functionally equivalent to fine-tuning the model on the current context \textit{at inference time}.

\textbf{Implication.} The global meta-parameters $\theta^*$ do not need to encode the conflict between Code and Medical knowledge. Instead, $\theta^*$ only needs to encode the \textit{universal rule}: ``If context involves Python syntax, update $\mathbf{S}$ to store code logic; if context involves Anatomy, update $\mathbf{S}$ to store biological relations''. Since this rule is consistent across domains, the Non-IID conflict in the parameter space is significantly alleviated.

Consequently, FedNL achieves \textbf{Zero-Shot Test-Time Adaptation}: even if the global model has never seen a specific local distribution during training, it can adapt to it during the first few tokens of inference, purely by executing the learned memory update rules. This property may improve robustness to certain forms of heterogeneity, especially when useful information can be absorbed from the test-time context.

%%%%%%%%%%%%%%%%%%%%%%%%%%%%%%%%%%%%%%%%%%%%%%%%%%%%%%%%%%%%%%%%%%%%%%%%%%%%%%%
\section{Experiments}
\label{sec:experiments}

We evaluate FedNL on two federated settings that stress different aspects of heterogeneity: a five-client Non-IID MMLU~\citep{hendryckstest2021} split for domain-specialized reasoning, and long-context NIAH tasks for sparse retrieval and streaming adaptation.

\subsection{Experimental Setup}
\label{subsec:setup}

\textbf{Setup Summary.} We evaluate FedNL on two federated settings: a five-client Non-IID MMLU split for domain-specialized short-context reasoning, and long-context NIAH tasks for multi-needle retrieval and streaming CE diagnostics. We additionally use PG-19 to isolate component-level effects in the ablation study. All experiments were conducted on 4 NVIDIA L20 48GB GPUs. The full data format, client partitions, and implementation details are provided in Appendix~\ref{app:data_format}.

\textbf{Baselines.} We compare FedNL with six representative federated methods spanning algorithmic and architectural axes. FedAvg~\citep{mcmahan2017communication} is the canonical FL baseline that averages local LoRA updates across clients. FedProx~\citep{li2020federated} augments FedAvg with a proximal term to mitigate client drift under heterogeneity. FedSSI~\citep{li2025fedssi} represents the current continual-FL regularization family, using synaptic-intelligence-style importance weights to preserve parameters across clients. FedALA~\citep{zhang2023fedala} personalizes the global model by locally calibrating aggregated weights on each client's data via a few SGD steps before evaluation. FFA-LoRA~\citep{sun2024ffalora} freezes the LoRA $A$ matrices at initialization and only averages the $B$ matrices across clients, reducing communication cost by half. Fed-Mamba is a backbone-comparison baseline that applies FedAvg to a Mamba-1.4B~\citep{gu2024mamba} state-space backbone, isolating the difference between SSM-style and Titans-style memory under federation.

\subsection{Federated Generalization on Non-IID MMLU}
\label{subsec:zero_shot_adaptation}
\label{subsec:mmlu_main}

We first study domain-level heterogeneity on MMLU. The benchmark is split into five clients, each corresponding to one super-category: Law/Ethics, Humanities, STEM, Math/CS, and Medical/Psychology. Each client fine-tunes on its own domain-specific training questions. The server then aggregates the client updates and redistributes the federated model back to all clients. Evaluation is performed on each client's held-out test questions from the same domain; these examples are unseen during training, and no gradient updates are performed at inference time.

\begin{figure}[H]
\centering
\includegraphics[width=0.98\linewidth]{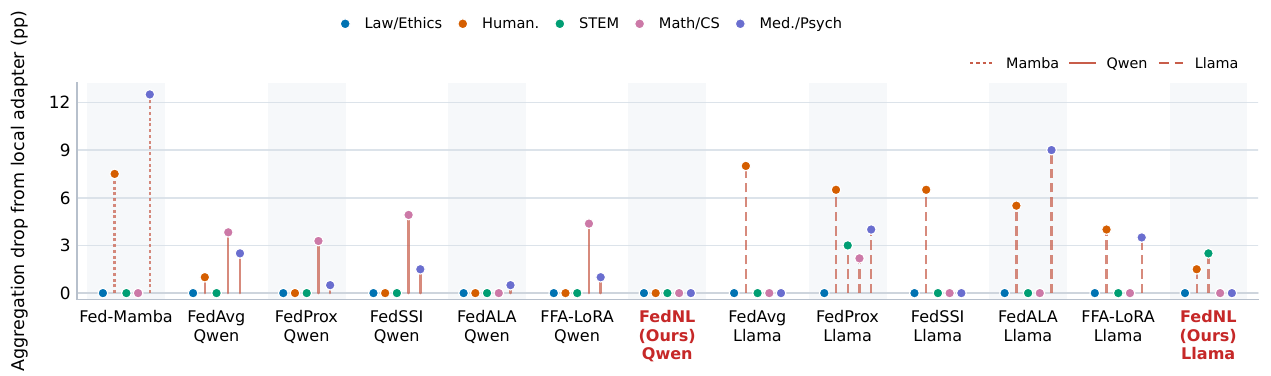}
\caption{Per-client MMLU aggregation drop from each client's locally fine-tuned adapter to the corresponding federated adapter. Lower is better: FedNL keeps the drop near zero across client domains, whereas static aggregation baselines show larger client-specific degradation under Non-IID shifts.}
\label{fig:mmlu_generalization_drop}
\end{figure}

Table~\ref{tab:mmlu_results} reports test accuracy on this five-client Non-IID partition, while Figure~\ref{fig:mmlu_generalization_drop} visualizes the client-level aggregation drop. On Qwen2.5-1.5B, FedNL obtains the highest average accuracy, $58.88\%$, slightly above FedSSI at $58.70\%$. The main gains appear on STEM and Math/CS, where FedNL improves over FedSSI by $+2.0$ and $+3.8$ percentage points, respectively. On the smaller Llama-3.2-1B backbone, FedNL reaches $42.64\%$, compared with $42.10\%$ for FedSSI, with the largest gain again on Math/CS. Fed-Mamba, which replaces the Titans-style memory with an SSM backbone, obtains $26.70\%$ average accuracy. These results suggest that memory-rule aggregation can be integrated into federated training without degrading short-context MMLU performance, while providing modest gains on the more shifted client domains.

\begin{table}[t]
\caption{Test accuracy (\%) on the five-client Non-IID MMLU partition. Each client trains on its own domain and is evaluated on held-out questions from that domain after federated aggregation. FedNL is evaluated on Titans-Qwen2.5-1.5B and Titans-Llama-3.2-1B against matched-backbone FL baselines; Fed-Mamba uses Mamba-1.4B as a non-Titans memory architecture.}
\label{tab:mmlu_results}
\centering
\setlength{\tabcolsep}{2.8pt}
\small
\begin{tabular}{>{\raggedright\arraybackslash}p{2.2cm}lcccccc}
\toprule
\textbf{Method} & \textbf{Backbone} & \textbf{Law/Eth} & \textbf{Human} & \textbf{STEM} & \textbf{Math/CS} & \textbf{Med/Psy} & \textbf{Avg} \\
\midrule
Fed-Mamba  & Mamba-1.4B   & 27.0 & 24.0 & 29.5 & 32.2 & 21.0 & 26.70 \\
\specialrule{0.08em}{0.6ex}{0.6ex}
FedAvg     & \multirow{5}{*}{Qwen2.5-1.5B} & 47.0 & 72.5 & 53.0 & 48.6 & 69.0 & 58.00 \\
FedProx    &                               & 47.0 & 73.0 & 52.0 & 49.2 & 69.0 & 58.00 \\
FedSSI     &                               & 48.5 & 74.0 & 53.0 & 48.1 & 70.0 & 58.70 \\
FedALA     &                               & 47.0 & 70.5 & 51.5 & 51.4 & 70.0 & 58.08 \\
FFA-LoRA   &                               & 47.5 & 73.0 & 54.5 & 47.5 & 68.0 & 58.11 \\
\specialrule{0.08em}{0.6ex}{0.6ex}
\cellcolor{red!10}\textbf{FedNL (Ours)} & \cellcolor{red!10}\textbf{Titans-Qwen2.5-1.5B} & \cellcolor{red!10}44.5 & \cellcolor{red!10}71.5 & \cellcolor{red!10}\textbf{55.0} & \cellcolor{red!10}\textbf{51.9} & \cellcolor{red!10}\textbf{71.5} & \cellcolor{red!10}\textbf{58.88} \\
\midrule
FedAvg     & \multirow{5}{*}{Llama-3.2-1B} & 34.0 & 45.0 & 38.0 & 37.7 & 51.5 & 41.20 \\
FedProx    &                               & 32.5 & 41.0 & 37.5 & 31.7 & 48.5 & 38.20 \\
FedSSI     &                               & 33.5 & 46.0 & 40.0 & 40.4 & 50.5 & 42.10 \\
FedALA     &                               & 33.5 & 42.5 & 38.5 & 40.4 & 48.0 & 40.59 \\
FFA-LoRA   &                               & 33.0 & 47.0 & 37.0 & 38.3 & 48.5 & 40.75 \\
\specialrule{0.08em}{0.6ex}{0.6ex}
\cellcolor{red!10}\textbf{FedNL (Ours)} & \cellcolor{red!10}\textbf{Titans-Llama-3.2-1B} & \cellcolor{red!10}33.5 & \cellcolor{red!10}43.5 & \cellcolor{red!10}38.5 & \cellcolor{red!10}\textbf{43.7} & \cellcolor{red!10}\textbf{54.0} & \cellcolor{red!10}\textbf{42.64} \\
\bottomrule
\end{tabular}
\end{table}

\subsection{Long-Tail Retrieval and Catastrophic Forgetting}
\label{subsec:long_tail}

We next evaluate long-tail retrieval with the NIAH suite under a seven-client Non-IID split. Each client corresponds to one retrieval template: MK-NIAH, MV-NIAH~\citep{hsieh2024ruler}, Passkey, UUID code, Name-date, Phrase code, or Counter state. Each client fine-tunes on its own template-specific training pool. The server then aggregates the client updates and redistributes the federated model back to all clients. Evaluation is performed on each client's held-out examples from the same retrieval template. All examples use multi-needle contexts at target depths from $1$K to $16$K tokens, so the task requires binding the queried key, rank, or state to the correct value rather than merely detecting that a needle-like value appeared in context. The full prompt construction and insertion rules are provided in Appendix~\ref{app:niah_data}.

\begin{figure}[t]
\centering
\newlength{\niahfigheight}
\setlength{\niahfigheight}{0.165\textheight}
\begin{minipage}[t]{0.515\linewidth}
\centering
\includegraphics[height=\niahfigheight]{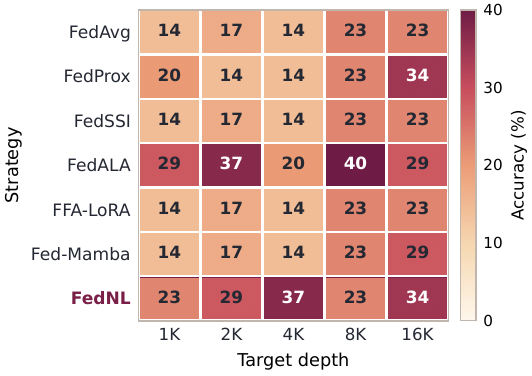}
\caption{NIAH accuracy by target insertion depth, averaged over the seven needle clients.}
\label{fig:niah_depth_heatmap}
\end{minipage}\hfill
\begin{minipage}[t]{0.445\linewidth}
\centering
\includegraphics[height=\niahfigheight]{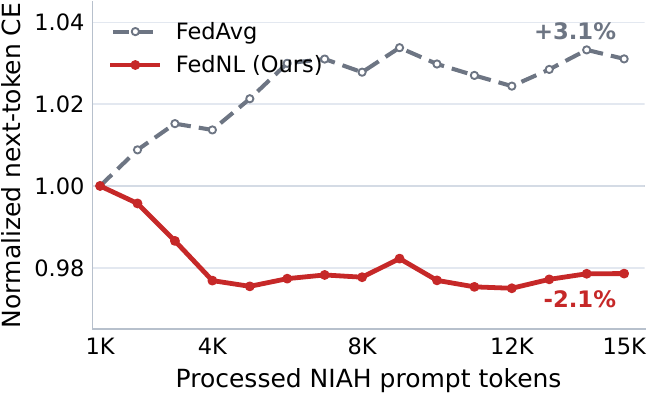}
\caption{16K NIAH streaming CE relative to each method's first 1K bin.}
\label{fig:niah_streaming_ce}
\end{minipage}
\end{figure}

Table~\ref{tab:niah_results} reports personalized accuracy on this seven-client NIAH partition, while Figure~\ref{fig:niah_depth_heatmap} averages the same evaluation over needle types at each target depth. FedNL obtains the highest average accuracy, $29.7\%$, compared with the strongest baseline FedALA at $28.6\%$. The largest gains appear on MK-NIAH, MV-NIAH, and UUID code, where FedNL reaches $32.0\%$, $40.0\%$, and $48.0\%$, respectively. The depth-stratified view shows that FedNL is strongest at $1$K--$4$K and remains tied for the best result at $16$K, while the harder $8$K setting narrows the gap across methods. To complement the accuracy view with a loss-based streaming diagnostic, Figure~\ref{fig:niah_streaming_ce} evaluates normalized next-token CE on 16K held-out NIAH prompts. FedAvg's CE increases by $3.1\%$ as the prompt unfolds, whereas FedNL decreases by $2.1\%$, indicating that the recurrent memory state continues to absorb useful context over long streams rather than accumulating uncertainty.

\begin{table}[H]
\caption{Per-client NIAH accuracy (\%) under the 7-client non-IID partition with multi-needle retrieval, averaged over target depths $1$K, $2$K, $4$K, $8$K, and $16$K (final round, personalized). FedAvg, FedProx, FedSSI, FedALA, and FFA-LoRA use the Llama-3.2-1B Transformer backbone; FedNL uses Titans-Llama-3.2-1B; Fed-Mamba uses Mamba-1.4B.}
\label{tab:niah_results}
\centering
\setlength{\tabcolsep}{3.2pt}
\footnotesize
\begin{tabular}{lcccccc>{\columncolor{red!10}}c}
\toprule
\textbf{Client} & \textbf{FedAvg} & \textbf{FedProx} & \textbf{FedSSI} & \textbf{Fed-Mamba} & \textbf{FedALA} & \textbf{FFA-LoRA} & \textbf{FedNL (Ours)} \\
\midrule
MK-NIAH       & 16.0 & 24.0 & 16.0 & 12.0 & 40.0 & 13.3 & 32.0 \\
MV-NIAH       & 24.0 & 16.0 & 24.0 & 20.0 & 40.0 & 20.0 & \textbf{40.0} \\
\midrule
Passkey       & \textbf{24.0} & \textbf{24.0} & \textbf{24.0} & 20.0 & 33.3 & 26.7 & \textbf{24.0} \\
UUID code     & 16.0 & 8.0  & 16.0 & 20.0 & 33.3 & 13.3 & \textbf{48.0} \\
Name-date     & 16.0 & 16.0 & 16.0 & 20.0 & 13.3 & 13.3 & \textbf{24.0} \\
Phrase code   & 20.0 & \textbf{32.0} & 20.0 & 28.0 & 20.0 & 13.3 & 28.0 \\
Counter state & 12.0 & \textbf{20.0} & 12.0 & 12.0 & 20.0 & 6.7  & 12.0 \\
\midrule
\textbf{Average} & 18.3 & 20.0 & 18.3 & 18.9 & 28.6 & 15.2 & \textbf{29.7} \\
\bottomrule
\end{tabular}
\end{table}

\subsection{Ablation Study}
\label{subsec:ablation}

Figure~\ref{fig:ablation} isolates the contribution of the main FedNL components. The full model keeps both the optimization-based Delta Rule and the Memory-as-Gate path while training LoRA adapters together with the memory parameters. We compare it with three controlled variants: \textbf{w/o Delta Rule}: replaces the Delta Rule with a Hebbian-style update, testing whether simple associative accumulation is sufficient; \textbf{w/o MaG}: removes the learned memory gate, forcing the model to rely on the memory path without the same fallback control; \textbf{w/o LoRA}: freezes the LoRA adapters and trains only the 32K memory parameters.
The Delta Rule is the most critical component: replacing it with a Hebbian update raises PPL from 29.82 to 1576.42, showing that simple accumulation cannot reliably correct noisy or overwritten memory values during streaming updates. Removing MaG increases PPL to 348.97 because the model loses a learned fallback that controls when to trust the memory path. Freezing LoRA gives 149.46 PPL, indicating that the small memory-rule parameters still need adapter-level alignment to make the recurrent state useful for language modeling.

\subsection{Communication and Resource Efficiency}
\label{subsec:efficiency}

A critical requirement for FL is efficiency. We analyze the inference memory footprint and the per-round communication cost.

\textbf{Inference Memory (VRAM).} Peak VRAM measurements compare memory usage across sequence lengths. Static-attention baselines grow with sequence length due to the KV cache and encounter Out-Of-Memory errors at 16k on resource-constrained accelerators. FedNL instead maintains a \textbf{constant $O(1)$ memory footprint} by storing only the fixed-size state $\mathbf{S}_t$, making long-context deployment more practical on edge devices.

\begin{figure}[t]
\centering
\begin{minipage}[H]{0.48\textwidth}
\centering
\includegraphics[width=\linewidth]{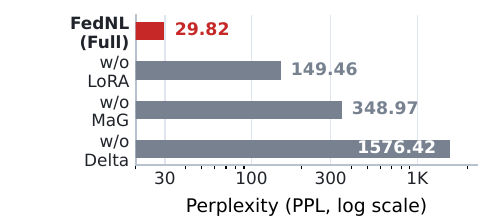}
\caption{PG-19 ablation (PPL, lower is better).}
\label{fig:ablation}
\end{minipage}
\hfill
\begin{minipage}[H]{0.48\textwidth}
\centering
\includegraphics[width=\linewidth]{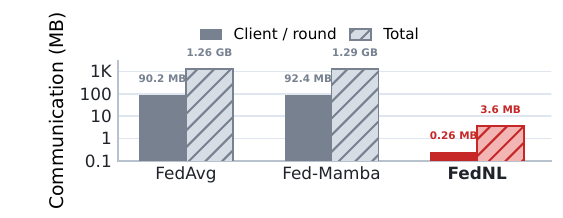}
\caption{Per-round communication on NIAH (Llama-3.2-1B, fp16).}
\label{fig:comm_cost}
\end{minipage}
\end{figure}

\textbf{Communication Efficiency.} Because FedNL aggregates only the memory-update meta-rules across clients --- a small fraction of the trainable parameter count --- the per-round client-to-server payload shrinks dramatically. On the NIAH 7-client setup (Llama-3.2-1B + Titans-Llama, $r{=}16$ LoRA), the effective memory rules amount to $\sim$$32{,}768$ parameters ($\sim$$0.26$\,MB at fp16), compared to $\sim$$11.3$\,M LoRA parameters ($\sim$$22.5$\,MB at fp16) for the FedAvg Transformer baseline. This is a $\mathbf{\sim}\bm{350\times}$ \textbf{reduction in per-round communication} (Figure~\ref{fig:comm_cost}). Aggregated over the full $7\!\times\!2$ training schedule, FedNL exchanges only $3.6$\,MB of cross-device traffic against $1.26$\,GB for FedAvg --- a property that is essential for deployment on bandwidth-constrained edge networks.

%%%%%%%%%%%%%%%%%%%%%%%%%%%%%%%%%%%%%%%%%%%%%%%%%%%%%%%%%%%%%%%%%%%%%%%%%%%%%%%
\section{Conclusion}
\label{sec:conclusion}
This paper proposes Federated Nested Learning (FedNL), a three-level nested optimization framework that redefines collaborative training. By theorizing FL as self-referential update rules, FedNL fundamentally addresses the Non-IID challenge. FedNL implements this theory through a Titans-based linear attention mechanism, enabling efficient zero-shot test-time adaptation. Empirical validation across Non-IID MMLU and long-context benchmarks demonstrates FedNL's stronger federated generalization, retrieval accuracy, and streaming CE behavior. This work establishes a new direction for FL, where models evolve from repository of knowledge into a paradigm of continuous, context-aware learning.

\section{Limitations}
\label{sec:limitations}
While our experiments demonstrate consistent gains at the 1B--1.5B scale, extending FedNL to larger foundation models remains an important next step to validate the generality of memory-rule aggregation. The empirical evaluation spans reasoning and retrieval tasks on MMLU and NIAH, and broadening the benchmark suite would further solidify the practical scope of the framework. 

% On the theoretical front, we formalize the three-level nested optimization structure, yet deriving rigorous convergence guarantees for the meta-learning loop presents an exciting avenue for future investigation.

\bibliography{reference}
\bibliographystyle{unsrtnat}

%%%%%%%%%%%%%%%%%%%%%%%%%%%%%%%%%%%%%%%%%%%%%%%%%%%%%%%%%%%%
\clearpage
\appendix

\section{Related Work}
\label{sec:related_work}
%Our work sits at the intersection of Continual Federated Learning, Memory-Augmented Neural Networks, and Test-Time Adaptation. Here we distinguish FedNL from prior arts in these domains.

\subsection{Continual Federated Learning}
Standard Federated Learning (FL) aggregates local updates to train a global model~\citep{mcmahan2017communication}. To handle statistical heterogeneity (Non-IID), methods like FedProx~\citep{li2020federated} and SCAFFOLD~\citep{karimireddy2020scaffold} introduce regularization or control variates. However, these methods assume a static data distribution over time.

Continual Federated Learning (CFL) addresses the scenario where clients face streaming tasks~\citep{yoon2021federated}. Existing approaches fall into two main categories:
\textbf{(1) Replay-based methods}~\citep{liu2020generative, qi2023better} store or generate past samples to rehearse old tasks. While effective, they fundamentally contradict the privacy-preserving ethos of FL and incur significant storage costs on edge devices.
\textbf{(2) Regularization-based methods} aim to constrain weight updates to protect important parameters. EWC~\citep{kirkpatrick2017overcoming} and Synaptic Intelligence (SI)~\citep{zenke2017continual} are classic examples. Recently, \textbf{FedSSI}~\citep{li2025fedssi} advanced this direction by introducing Personalized Surrogate Models (PSM) to calibrate local SI regularization with global information, achieving state-of-the-art performance in preventing catastrophic forgetting.

\textbf{Limitations of Current CFL:} Despite their sophistication, methods ranging from FedAvg to FedSSI share a common premise: they treat the global model as a container of \textit{static knowledge}. They aim to find a set of weights $\theta^*$ that creates a compromise between conflicting tasks. In contrast, FedNL fundamentally departs from this ``static weight'' paradigm. Instead of regularizing weights to prevent them from changing, we design the model to \textit{actively change} its internal state ($\mathbf{S}_t$) during inference, enabling it to embrace heterogeneity rather than compromise with it.

\subsection{Nested Learning and Neural Memory}
The concept of \textbf{Nested Learning (NL)}~\citep{behrouz2025nested} posits that intelligent systems should be modeled as hierarchies of optimization loops operating at different frequencies. This framework unifies meta-learning~\citep{finn2017model} and in-context learning under a single theoretical umbrella. A practical realization of NL is the \textbf{Titans} architecture~\citep{behrouz2024titans}, which utilizes a linearized attention mechanism equipped with a memory module. Unlike standard Recurrent Neural Networks (RNNs) or State Space Models (Mamba)~\citep{gu2024mamba} that use fixed heuristic updates, Titans updates its memory via the \textbf{Delta Rule} --- mechanistically equivalent to an online gradient descent step.
FedNL is the first work to apply the NL perspective to Federated Learning. We reinterpret the client's inference process as the ``inner-loop'' optimization defined in NL, and the federated aggregation as the ``outer-loop'' meta-learning. This allows us to decouple the \textit{memory content} (local, private, transient) from the \textit{memory update rules} (global, shared, persistent).  

\subsection{Test-Time Training and Adaptation}
Test-Time Training (TTT)~\citep{sun2020test, wang2020tent} refers to the paradigm of updating model parameters during inference to adapt to distribution shifts. Recent advancements, such as TTT-Linear~\citep{sun2024learning}, bake this optimization directly into the forward pass of sequence models. FedNL can be viewed as a \textbf{Federated Collaborative TTT} framework. While standard TTT focuses on adapting a single isolated model, FedNL aggregates the experience of multiple clients to learn \textit{how to adapt} efficiently. By learning the optimal meta-parameters $\theta$ (projections and gating), FedNL ensures that the test-time adaptation (via Delta Rule) is robust and converges rapidly on unseen Non-IID domains. This effectively solves the ``cold-start'' problem often faced by TTT methods in zero-shot scenarios.

\section{Derivations and Implementation Details}
\label{app:derivations}

In this appendix, we provide the detailed mathematical derivations supporting the theoretical framework of Federated Nested Learning (FedNL). Specifically, we analyze the gradient flow through the dynamic memory states to validate the meta-learning interpretation of our method (Level 1 Loop). We also detail the efficient chunk-wise parallel implementation of the Delta Rule used in the Inner Loop (Level 2).

\subsection{Gradient Flow Analysis: Optimizing the Learning Rule}
\label{app:gradient_flow}

In Section \ref{sec:preliminaries}, we defined the local training objective for a client $k$ as finding the optimal meta-parameters $\theta$ that minimize the cumulative prediction loss over a sequence $x_{1:T}$. The loss is given by:
\begin{equation}
    \mathcal{J}(\theta) = \sum_{t=1}^T \ell \left( f(x_t; \mathbf{S}_{t-1}, \theta), x_{t+1} \right),
\end{equation}
where $\mathbf{S}_t$ evolves according to the Delta Rule (Eq. \ref{eq:delta_rule}):
\begin{equation}
    \mathbf{S}_t = \mathbf{S}_{t-1} + \beta_t (\mathbf{v}_t - \mathbf{S}_{t-1}\mathbf{k}_t)\mathbf{k}_t^\top.
    \label{eq:app_delta_rule}
\end{equation}
Here, $\mathbf{k}_t, \mathbf{v}_t, \beta_t$ are functions of the input $x_t$ and parameters $\theta$ (specifically the LoRA adapters and gating networks).

To update $\theta$ using Gradient Descent (Level 1 Loop), we require the total derivative $\frac{d \mathcal{J}}{d \theta}$. Applying the chain rule through time (BPTT), the gradient at step $t$ depends on the state $\mathbf{S}_{t-1}$, which in turn depends on $\theta$ through all previous timesteps.

The total gradient can be expanded as:
\begin{equation}
    \frac{d \mathcal{J}}{d \theta} = \sum_{t=1}^T \left( 
    \underbrace{\frac{\partial \ell_t}{\partial \theta}}_{\text{Direct}} + 
    \underbrace{\frac{\partial \ell_t}{\partial \mathbf{S}_{t-1}} \cdot \frac{d \mathbf{S}_{t-1}}{d \theta}}_{\text{Recursive}} 
    \right).
\end{equation}

The Direct Term captures how $\theta$ affects the immediate prediction (e.g., through the output projection layer). The Recursive Term captures the ``meta-learning'' signal: how $\theta$ influences the \textit{construction} of the memory.

We can expand the recursive state derivative $\frac{d \mathbf{S}_t}{d \theta}$ using Eq. (\ref{eq:app_delta_rule}):
\begin{equation}
    \frac{d \mathbf{S}_t}{d \theta} = \frac{\partial \mathbf{S}_t}{\partial \mathbf{S}_{t-1}} \frac{d \mathbf{S}_{t-1}}{d \theta} + \frac{\partial \mathbf{S}_t}{\partial \theta}\bigg|_{\mathbf{S}_{t-1} \text{ fixed}}.
\end{equation}

\textbf{1. The Transition Jacobian ($\frac{\partial \mathbf{S}_t}{\partial \mathbf{S}_{t-1}}$):}
Differentiating Eq. (\ref{eq:app_delta_rule}) w.r.t $\mathbf{S}_{t-1}$:
\begin{equation}
    \frac{\partial \mathbf{S}_t}{\partial \mathbf{S}_{t-1}} = \mathbf{I} - \beta_t \mathbf{k}_t \mathbf{k}_t^\top.
\end{equation}
This term acts as a "forgetting gate" or contraction map. It determines how much of the gradient flows back to previous memories. In FedNL, $\theta$ learns to generate $\mathbf{k}_t$ and $\beta_t$ such that this Jacobian preserves gradients for relevant long-term dependencies while dampening noise.

\textbf{2. The Update Jacobian ($\frac{\partial \mathbf{S}_t}{\partial \theta}\big|_{\text{direct}}$):}
This term represents how a change in $\theta$ alters the \textit{content} written into memory at step $t$. Since $\mathbf{k}_t, \mathbf{v}_t, \beta_t$ are functions of $\theta$:
\begin{equation}
    \frac{\partial \mathbf{S}_t}{\partial \theta}\bigg|_{\text{direct}} \approx \beta_t \left( \frac{\partial \mathbf{v}_t}{\partial \theta}\mathbf{k}_t^\top + \mathbf{v}_t\frac{\partial \mathbf{k}_t^\top}{\partial \theta} \right) + (\dots).
\end{equation}
By optimizing this term, FedNL explicitly trains the projection matrices (e.g., LoRA $A, B$) to produce keys and values that maximize the utility of the resulting memory trace.

\textbf{Conclusion:} The gradient descent update on $\theta$ in the Local Loop effectively solves a meta-optimization problem: \textit{"Find the projection rules $\theta$ such that executing the Delta Rule (Inner Loop) yields the sequence of states $\mathbf{S}_{0:T}$ that minimizes prediction error."}

\subsection{Efficient Chunk-wise Parallelization}
\label{app:parallelization}

While the Delta Rule (Eq. \ref{eq:app_delta_rule}) is recurrent and seemingly sequential ($O(T)$), we leverage the properties of linear recurrence to parallelize computation, making FedNL feasible for edge devices.

We divide the input sequence of length $T$ into chunks of size $C$ (e.g., $C=128$). The computation is decomposed into \textit{Intra-Chunk} (parallel) and \textit{Inter-Chunk} (recurrent) operations.

\textbf{Matrix Formulation of Delta Rule.}
Eq. (\ref{eq:app_delta_rule}) can be rewritten as a linear recurrence:
\begin{equation}
    \mathbf{S}_t = \mathbf{S}_{t-1} \mathbf{W}_t + \mathbf{U}_t,
\end{equation}
where $\mathbf{W}_t = (\mathbf{I} - \beta_t \mathbf{k}_t \mathbf{k}_t^\top)$ is the decay matrix and $\mathbf{U}_t = \beta_t \mathbf{v}_t \mathbf{k}_t^\top$ is the update term.

\textbf{1. Intra-Chunk Computation (Parallel):}
For a chunk $b$ spanning time steps $i$ to $j$, we can compute the aggregate transition matrix $\mathbf{W}_{b}$ and aggregate update $\mathbf{U}_{b}$ in parallel.
Because $\mathbf{W}_t$ is a rank-1 perturbation of the identity, the cumulative product over the chunk can be computed efficiently using the WY representation \citep{sun2024learning} or parallel associative scans.
Specifically, we compute the local memory states $\tilde{\mathbf{S}}_{t}$ within the chunk assuming a zero initial state ($\mathbf{S}_{i-1} = \mathbf{0}$). This can be implemented via standard causal self-attention masks within the chunk:
% \begin{equation}
%     \text{ChunkOutput}_b = \text{CausalDotProduct}(\mathbf{Q}_b, \mathbf{K}_b, \mathbf{V}_b, \bm{\beta}_b).
% \end{equation}
\begin{equation}
    \text{ChunkOutput}_b = \text{CausalDotProduct}(\mathbf{Q}_b, \mathbf{K}_b, \mathbf{V}_b, {\beta}_b).
\end{equation}

\textbf{2. Inter-Chunk Recurrence:}
Once the aggregate effect of each chunk is computed, we update the boundary states $\mathbf{S}_{b}$ sequentially:
\begin{equation}
    \mathbf{S}_{b} = \mathbf{S}_{b-1} \mathbf{W}_{chunk\_b} + \mathbf{U}_{chunk\_b}.
\end{equation}
Since the number of chunks $T/C$ is small, this sequential step is negligible.

\textbf{3. Final Output:}
The query $\mathbf{q}_t$ at any time $t$ inside chunk $b$ interacts with both the intra-chunk local memory and the passed-down inter-chunk memory:
\begin{equation}
    \mathbf{o}_t = \underbrace{(\mathbf{S}_{b-1} \prod_{\tau=i}^t \mathbf{W}_\tau) \mathbf{q}_t}_{\text{Long-term History}} + \underbrace{\tilde{\mathbf{S}}_t \mathbf{q}_t}_{\text{Local Context}}.
\end{equation}

\textbf{Efficiency Analysis.}
For training at Level 1, this chunk-wise formulation allows us to train on long sequences (e.g., 4k tokens) with high GPU utilization, as the heavy lifting is done by parallel matrix multiplications (Tensor Cores).
For inference at Level 2, token-by-token generation reverts to the $O(1)$ recurrent form (Eq.~\ref{eq:app_delta_rule}), ensuring constant memory usage and low latency on edge devices.

\section{Benchmark Data Format}
\label{app:data_format}

This appendix describes the concrete example format used in the two federated benchmarks. In both cases, the client identity is tied to the data domain rather than sampled from a global mixture.

\subsection{Experimental Setup Details}
\label{app:experimental_setup}

We construct two scenarios to simulate extreme heterogeneity and long-tail distributions. The first is a Non-IID MMLU split, where the MMLU benchmark~\citep{hendrycks2020measuring} is partitioned into five disjoint super-categories, one per client: Law/Ethics, Humanities, STEM, Math/CS, and Medical/Psychology. Each client holds approximately 2{,}000 training and up to 200 test questions drawn from its assigned domain, producing a setting where no client observes the global subject mixture. The second scenario combines the Needle In A Haystack benchmark and PG-19 to evaluate long-tail recall and language modeling over long horizons, up to 16k tokens.

For MMLU, we evaluate FedNL on two backbone scales, Qwen2.5-1.5B and Llama-3.2-1B, against matched-backbone FL baselines; Fed-Mamba uses Mamba-1.4B as a non-Titans memory comparison. Long-context experiments use Llama-3.2-1B and Titans-Llama-3.2-1B respectively. Trainable parameters are restricted to LoRA adapters and, for FedNL, the LiZAttention~\citep{furfaro2025tptt} memory parameters. For MMLU we use LoRA rank $r{=}32$, $\alpha{=}64$ at learning rate $5\!\times\!10^{-5}$; for the NIAH and PG-19 experiments we use $r{=}16$, $\alpha{=}32$ at learning rate $3\!\times\!10^{-4}$. All methods train for one local epoch per round. The NIAH experiments run for two federated rounds and report the final personalized accuracy.

\subsection{MMLU Federated Split}
\label{app:mmlu_data}

Each MMLU example is a four-choice multiple-choice question with a question stem, four answer options, a subject label, and a zero-indexed gold option. The gold index 0--3 corresponds to choices A--D. During loading, the row is converted into the prompt
\begin{quote}
\small
\texttt{Question: <question>}\\
\texttt{A. <choice 0>}\\
\texttt{B. <choice 1>}\\
\texttt{C. <choice 2>}\\
\texttt{D. <choice 3>}\\
\texttt{Answer:}
\end{quote}
Local training appends the gold answer letter and applies the loss to that answer token. Evaluation uses the same prompt and scores the first generated answer letter.

\begin{table}[h]
\caption{MMLU client partition used in the federated experiments. The train/test columns report the examples selected by the loader after shuffling and capping each client at at most 2{,}000 training and 200 test rows.}
\label{tab:app_mmlu_split}
\centering
\scriptsize
\setlength{\tabcolsep}{3pt}
\begin{tabular}{p{0.19\linewidth}p{0.55\linewidth}cc}
\toprule
\textbf{Super-category} & \textbf{MMLU subjects} & \textbf{Train} & \textbf{Test} \\
\midrule
Law/Ethics &
Professional law, moral scenarios, moral disputes, logical fallacies, formal logic, international law, jurisprudence, business ethics & 2000 & 200 \\
Humanities &
Miscellaneous, history, politics, economics, geography, sociology, philosophy, religion, management, marketing, public relations, security studies, global facts & 2000 & 200 \\
STEM &
Elementary mathematics, biology, chemistry, physics, astronomy, anatomy, electrical engineering & 2000 & 200 \\
Math/CS &
High-school mathematics, high-school statistics, machine learning, college mathematics, high-school computer science, college computer science, computer security, abstract algebra & 915 & 183 \\
Medical/Psychology &
Professional psychology, high-school psychology, virology, nutrition, professional medicine, clinical knowledge, human aging, college medicine, human sexuality, medical genetics & 2000 & 200 \\
\bottomrule
\end{tabular}
\end{table}

\subsection{NIAH Federated Split}
\label{app:niah_data}

The main NIAH setting uses seven clients: Passkey, UUID code, Name-date, Phrase code, Counter state, MK-NIAH, and MV-NIAH. Each client contains 750 training examples, a 15-example held-out test split balanced over target depths \(\{1024, 2048, 4096\}\), and an additional long-depth test split at \(\{8192, 16384\}\). Each example records the target depth, insertion positions, full prompt, four answer candidates, gold answer, correct answer letter, task metadata, and the inserted needle events.

The prompt is built by sampling a slice of WikiText-103 filler, inserting several needle events at controlled depth fractions, and appending a four-choice question:
\begin{quote}
\small
\texttt{<filler prefix> <event 1> <filler> ... <event m> <filler suffix>}\\
\texttt{Question: <retrieval question>}\\
\texttt{A) <candidate 1>}\\
\texttt{B) <candidate 2>}\\
\texttt{C) <candidate 3>}\\
\texttt{D) <candidate 4>}\\
\texttt{Answer:}
\end{quote}
The correct candidate is randomly assigned to one of A--D. Distractors are hard negatives from the same haystack whenever possible, so selecting a value that appeared in context is insufficient unless it is bound to the queried key, rank, or state.

\begin{table}[h]
\caption{Needle templates in the seven-client NIAH split. Each ordinary retrieval client inserts four target-like events per haystack; the question selects one event and uses the other observed values as distractors.}
\label{tab:app_niah_needles}
\centering
\scriptsize
\setlength{\tabcolsep}{3pt}
\begin{tabular}{p{0.16\linewidth}p{0.45\linewidth}p{0.27\linewidth}}
\toprule
\textbf{Client} & \textbf{Inserted event form} & \textbf{Question target} \\
\midrule
Passkey & Officer [name]'s security passkey is [7-digit number]. & Retrieve the passkey for a named officer. \\
UUID code & Operational access code for device [id] is [three-part code]. & Retrieve the code for a specified device. \\
Name-date & Officer [name] filed the registration report on [date]. & Retrieve the filing date for a specified officer. \\
Phrase code & Mission [name]'s activation codeword is [phrase-code]. & Retrieve the codeword for a specified mission. \\
Counter state & State base [state] followed by several state increment or state decrement events. & Compute the final state modulo 8. \\
MK-NIAH & The [key] secret is [4-digit value], repeated for multiple keys. & Retrieve the secret associated with the queried key. \\
MV-NIAH & Codeword 1: [value], Codeword 2: [value], etc. & Retrieve the value at the queried rank. \\
\bottomrule
\end{tabular}
\end{table}

For clients with four events, events are placed at the first four canonical depth fractions \(\{0.10, 0.30, 0.50, 0.70\}\) within the filler budget. Stateful clients such as counter state can contain more events; in that case event positions are spread approximately uniformly across the haystack. This construction makes all seven clients multi-needle retrieval tasks rather than single-needle lookup tasks.

\section{Broader Impacts}

FedNL aims to improve federated learning for language models under heterogeneous and long-context client data. Its potential positive impacts include enabling more adaptive on-device or edge language models, reducing the need to centralize raw user data, and lowering communication costs by exchanging compact memory-update rules rather than full model parameters. These properties may make privacy-preserving and bandwidth-efficient collaborative training more accessible in settings such as personalized assistants, domain-specific reasoning tools, and resource-constrained deployments.

At the same time, FedNL inherits several risks associated with adaptive language models and federated learning. First, improved test-time adaptation may make models more effective in benign applications, but it could also improve the capability of systems used for harmful purposes, such as generating misleading content, automating social engineering, or adapting to user-specific contexts in manipulative ways. Second, although FedNL keeps raw data and transient memory states local, federated updates can still carry privacy risks through model-update leakage or membership-inference attacks. Therefore, practical deployments should consider standard privacy protections such as secure aggregation, differential privacy, careful logging policies, and auditing of communicated updates.

Third, heterogeneous client distributions can create fairness and reliability concerns. A model that adapts strongly to local context may perform unevenly across domains, dialects, demographic groups, or low-resource settings, especially when some client distributions are underrepresented during federated training. In sensitive applications such as medicine, law, or education, incorrect test-time adaptation could lead to misleading or harmful outputs even when the system is used as intended. Deployments should therefore include domain-specific evaluation, uncertainty monitoring, human oversight, and safeguards against overconfident predictions.

Finally, FedNL is a methodological contribution rather than a deployed system. We do not release user data or propose an application-specific decision-making pipeline. Nevertheless, because the method can improve efficient adaptation of language models, downstream uses should be evaluated for privacy, security, fairness, and misuse risks before deployment.

\section{Asset Licenses and Terms of Use}
\label{app:asset_licenses}

Our experiments use existing pretrained model backbones, benchmark datasets, and software components. We cite the original sources for all assets used in the paper and use them only for research purposes in accordance with their respective licenses and terms of use.

The pretrained language-model backbones include Llama-3.2-1B, Qwen2.5-1.5B, and Mamba-1.4B. The federated and long-context experiments use public benchmark datasets including MMLU, NIAH/RULER-style retrieval tasks, PG-19, and WikiText-103 filler text for prompt construction. The implementation further builds on publicly described components such as LoRA, Titans-style memory, and LiZAttention. We do not claim ownership over these assets.

%%%%%%%%%%%%%%%%%%%%%%%%%%%%%%%%%%%%%%%%%%%%%%%%%%%%%%%%%%%%

% \newpage
% \input{checklist.tex}

\end{document}